\documentclass{article}

\PassOptionsToPackage{numbers, compress}{natbib}

\usepackage[final]{nips_2018}

\usepackage[utf8]{inputenc} 
\usepackage[T1]{fontenc}    
\usepackage{hyperref}       
\usepackage{url}            
\usepackage{booktabs}       
\usepackage{amsfonts}       
\usepackage{nicefrac}       
\usepackage{microtype}      

\usepackage{multirow}
\usepackage{amsmath}
\usepackage{lipsum}
\usepackage{graphicx}
\usepackage{algorithm}
\usepackage{algpseudocode}
\usepackage{tabularx}
\usepackage{booktabs}
\usepackage{amssymb}
\usepackage{multirow}
\newtheorem{theorem}{Theorem}

\usepackage[T1]{fontenc}
\usepackage{aecompl}
\usepackage[font=small,labelfont=bf]{caption}

\title{Structured Pruning for Efficient ConvNets via Incremental Regularization}

\author{
Huan Wang, Qiming Zhang, Yuehai Wang, Haoji Hu\thanks{Corresponding author} \\
Zhejiang University \\
\tt \small\{huanw, qmzhang, wyuehai, haoji\_hu\}@zju.edu.cn
}

\begin{document}

\maketitle

\begin{abstract}
Parameter pruning is a promising approach for CNN compression and acceleration by eliminating redundant model parameters with tolerable performance loss. Despite its effectiveness, existing regularization-based parameter pruning methods usually drive weights towards zero with \emph{large and constant} regularization factors, which neglects the fact that the expressiveness of CNNs is fragile and needs a more gentle way of regularization for the networks to adapt during pruning. To solve this problem, we propose a new regularization-based pruning method (named \emph{IncReg}) to incrementally assign different regularization factors to different weight groups based on their relative importance, whose effectiveness is proved on popular CNNs compared with state-of-the-art methods.
\end{abstract}

\vspace{-1.5em}
\section{Introduction}
\vspace{-0.5em}
Recently, deep Convolutional Neural Networks~(CNNs) have made a remarkable success in computer vision tasks by leveraging large-scale networks learning from big amount of data. However, CNNs usually lead to massive computation and storage consumption, thus hindering their deployment on mobile and embedded devices. To solve this problem, many research works focus on compressing the scale of CNNs. Parameter pruning is a promising approach for CNN compression and acceleration, which aims at eliminating redundant model parameters at tolerable performance loss. To avoid hardware-unfriendly irregular sparsity, structured pruning is proposed for CNN acceleration~\cite{AnwSun16,SzeCheYanEme17}. In the \verb+im2col+ implementation~\cite{ChePurSim06,CheWooVan14} of convolution, weight tensors are expanded into matrices, so there are generally two kinds of structured sparsity, i.e., row sparsity (or filter-wise sparsity) and column sparsity (or shape-wise sparsity)~\cite{WenWuWan16,Wang2018Structured}.

There are mainly two categories of structured pruning. One is importance-based methods, which prune weights in groups based on some established importance criteria~\cite{LiKadDurEtAl17,MolTyrKar17,Wang2018Structured}. The other is regularization-based methods, which add group regularization terms to learn structured sparsity~\cite{WenWuWan16,VadLem16,He2017Channel}. Existing group regularization approaches mainly focus on the regularization form (e.g., Group LASSO~\cite{Yuan2006Model}) to learn structured sparsity, while ignoring the influence of regularization factor. In particular, they tend to use a \emph{large and constant} regularization factor for all weight groups in the network~\cite{WenWuWan16,VadLem16}, which has two problems. Firstly, this `one-size-fit-all' regularization scheme has a hidden assumption that all weights in different groups are equally important, which however does not hold true, since weights with larger magnitude tend to be more important than those with smaller magnitude. Secondly, few works have noticed that the expressiveness of CNNs is so fragile~\cite{yosinski2014transferable} during pruning that it cannot withstand a large penalty term from beginning, especially for large pruning ratios and compact networks (like ResNet~\cite{HeZhaRenSun16}). AFP~\cite{DinDinHanTan18} was proposed to solve the first problem, while ignored the second one. In this paper, we propose a new regularization-based method named IncReg to \emph{incrementally} learn structured sparsity.

Apart from pruning, there are also other kinds of methods for CNN acceleration, such as low-rank decomposition~\cite{LebYarRakOseLem16,ZhaZouHeSun16}, quantization~\cite{CheWilTyrWeiChe15,CouBen16,LinCouMemBen16,RasOrdRedFar16}, knowledge distillation~\cite{Hinton2015Distilling} and architecture re-design~\cite{IanMosAsh16,Howard2017MobileNets,Zhang2017ShuffleNet,zhong2018shift}. Our method is orthogonal to these methods.

\vspace{-0.8em}
\section{The Proposed Method}
\vspace{-0.7em}
Given a conv kernel, modeled by a 4-D tensor~$\mathbf{W}^{(l)} \in \mathbb{R}^{N^{(l)} \times C^{(l)} \times H^{(l)} \times W^{(l)}}$, where~$N^{(l)}$, $C^{(l)}$, $H^{(l)}$ and~$W^{(l)}$ are the dimension of the~$l$th ($1 \leq l \leq L$) weight tensor along the axis of filter, channel, height and width, respectively, our proposed objective function for regularization can be formulated as: $E(\mathbf{W}) = L(\mathbf{W}) + \frac{\lambda}{2}R(\mathbf{W}) + \sum_{l=1}^L \sum_{g=1}^{G^{(l)}} \frac{\lambda_g^{(l)}}{2} R(\mathbf{W}^{(l)}_g)$, where~$\mathbf{W}$ denotes the collection of all weights in the CNN; $L(\mathbf{W})$ is the loss function for prediction; $R(\mathbf{W})$ is non-structured regularization on every weight, i.e., weight decay in this paper; $R(\mathbf{W}^{(l)}_g)$ is the structured sparsity regularization term on group $g$ of layer $l$ and $G^{(l)}$ is the number of weight groups in layer $l$. In~\cite{VadLem16,WenWuWan16}, the authors used the same~$\lambda_g$ for all groups and adopted Group LASSO~\cite{Yuan2006Model} for~$R(\mathbf{W}^{(l)}_g)$. In this work, since we emphasize the key problem of group regularization lies in the regularization factor rather than the regularization form, we use the most common regularization form weight decay, for~$R(\mathbf{W}^{(l)}_g)$, but we vary the regularization factors~$\lambda_g$ for different weight groups and at different iterations. Especially, we propose a theorem (Theorem~\ref{theorem:main} and its proof in Appendix) to show that we can gradually compress the magnitude of a parameter by adjusting its regularization factor.

Our method prunes all the conv layers simultaneously and independently. For simplicity, we omit the layer notation $l$ for following description. All the $\lambda_g$'s are initialized to zero. At each iteration, $\lambda_g$ is increased by $\lambda_g^{new} = \max(\lambda_g + \Delta \lambda_g, 0)$. Like AFP~\cite{DinDinHanTan18}, we agree that unimportant weights should be punished more, so we propose a decreasing piece-wise linear punishment function (Eqn.\ref{eqn:punish_function}, Fig.\ref{fig:punish_function}) to determine $\Delta \lambda_g$. Note that the regularization increment is negative (i.e., reward actually) when ranking is above the threshold ranking $RG$, since above-the-threshold means these weights are expected to stay in the end. Regularization on these important weights is not only unnecessary but also very harmful via our experimental confirmation.
\begin{equation}
\Delta\lambda_g(r) = \left\{
    \begin{aligned}
              & -\frac{A}{RG}r+A            &  \text{\emph{ if} }  r \leq RG\\
              &-\frac{A}{G(1-R)-1}(r-RG)  & \text{\emph{ if} } r >  RG
    \end{aligned}
    \right.
\label{eqn:punish_function}
\end{equation}
where $R$ is the pre-assigned pruning ratio for a layer, $G$ is the number of weight groups, $A$ is a hyper-parameter in our method to describe the maximum penalty increment (set to half of the original weight decay in default), $r$ is the ranking obtained by sorting in ascending order based on a proposed importance criterion, which is essentially an averaged ranking over time, defined as $\frac{1}{N}\sum_{n=1}^{N}r_n$, where $r_n$ is the ranking by $L_1$-norm at $n$th iteration, $N$ is the number of passed iterations. This averaging is adopted as smoothing for a more stable pruning process.

As training proceeds, the regularization factors of different weight groups increase gradually, which will push the weights towards zero little by little. When the magnitude of a weight group is lower than some threshold ($10^{-6}$), the weights are permanently removed from the network, thus leading to increased structured sparsity. When the sparsity of a layer reaches its pre-assigned pruning ratio $R$, that layer automatically stops structured regularization. Finally, when all conv layers reach their pre-assigned pruning ratios, pruning is over, followed by a retraining process to regain accuracy.



\vspace{-0.8em}
\section{Experiments}
\vspace{-0.7em}
\subsection{Analysis with ConvNet on CIFAR-10}
We firstly compare our proposed IncReg with other two group regularization methods, i.e., SSL~\cite{WenWuWan16} and AFP~\cite{DinDinHanTan18}, with ConvNet~\cite{KriSutHin12} on CIFAR-10~\cite{KriHin09}, where both row sparsity and column sparsity are explored. Caffe~\cite{JiaSheDonEtAl14} is used for all of our experiments. Experimental results are shown in Tab.\ref{tab:result_ssl_afp_ours}. We can see that IncReg consistently achieves higher speedups and accuracies than the other two constant regularization schemes. Notably, even though AFP achieves similar performance as our method under relatively small speedup (about $4.5\times$), when the speedup ratio is large (about $8\times\sim10\times$), our method outperforms AFP by a large margin. We argue that this is because the incremental way of regularization gives the network more time to adapt during pruning, which is especially important in face of large pruning ratios.

\begin{minipage}{\textwidth}
    \begin{minipage}[b]{0.40\textwidth}
      \centering
      \includegraphics[width=1.0\textwidth]{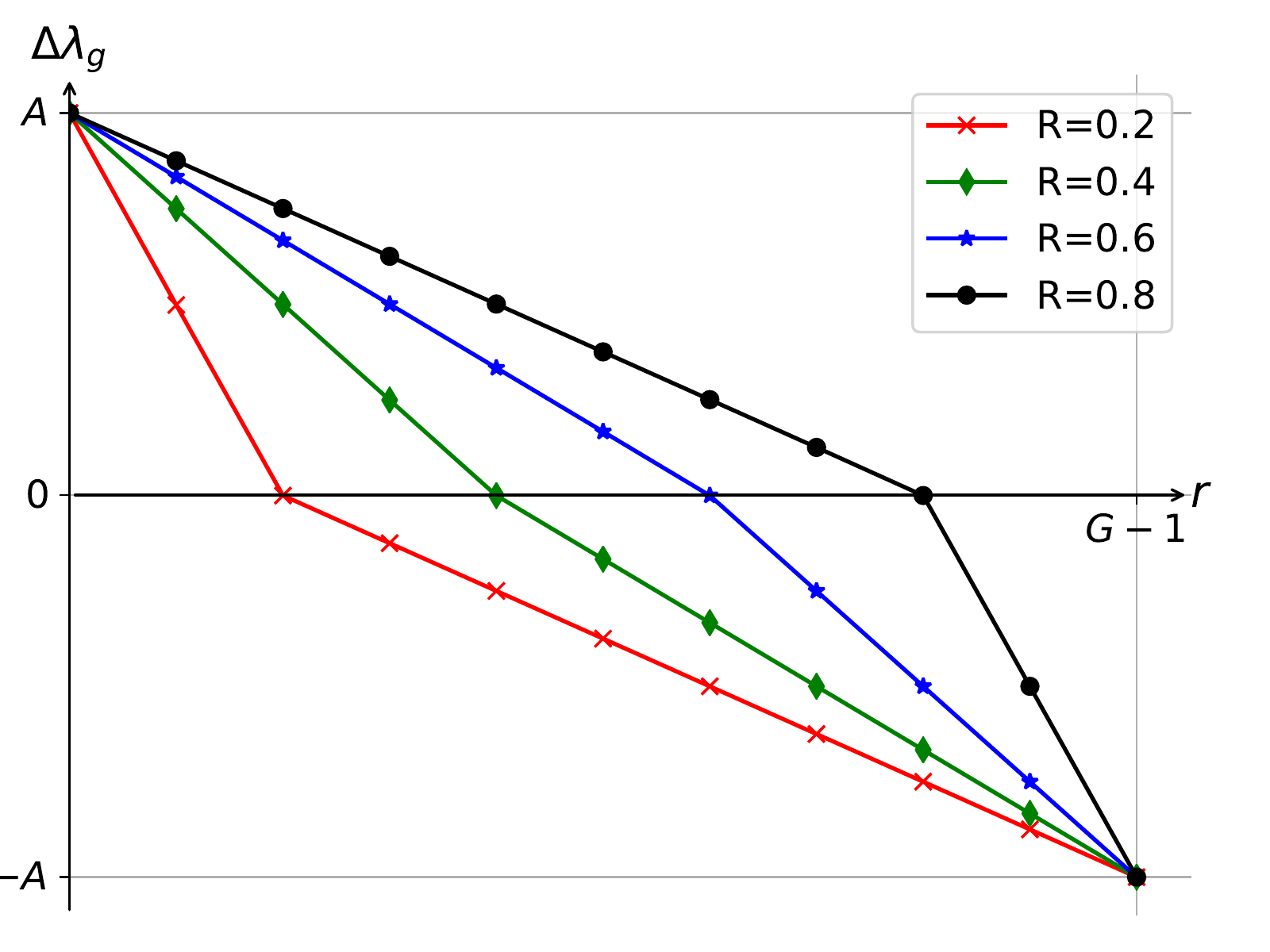}
      \captionof{figure}{The illustration of our proposed punishment function of $\Delta \lambda_g$.}
      \label{fig:punish_function}
    \end{minipage}
    \hfill
    \begin{minipage}[b]{0.59\textwidth}
      \centering
      \begin{tabular}{l p{1cm}<{\centering} p{1cm}<{\centering} p{1cm}<{\centering} p{1cm}<{\centering}}
         \toprule
         \multirow{2}*{Method}  & \multicolumn{2}{c}{Row sparsity} & \multicolumn{2}{c}{Column sparsity} \\
               \cline{2-5}
                   & speedup & accuracy & speedup & accuracy \\
         \hline
         SSL               & $3.6 \times$    & $77.3$          & $3.1\times$    & $78.6$  \\
         AFP~(our impl.)   & $4.1\times$     & $77.7$           & $4.5\times$          & $81.0$ \\
         Ours              & $\mathbf{4.1\times}$           & $\mathbf{79.2}$ & $\mathbf{4.6\times}$   & $\mathbf{81.2}$ \\
         \hline
         SSL               & $9.7\times$      & $73.0$              & $10.0\times$    & $75.2$ \\
         AFP~(our impl.)   & $9.9\times$      & $73.4$              & $8.4\times$          & $77.5$ \\
         Ours              & $\mathbf{9.9\times}$      & $\mathbf{76.0}$              & $\mathbf{10.0\times}$    & $\mathbf{78.7}$ \\                  
       \bottomrule
      \end{tabular}
      \captionof{table}{Comparison of our method with SSL~\cite{WenWuWan16} and AFP~\cite{DinDinHanTan18} with ConvNet on CIFAR-10.}
      \label{tab:result_ssl_afp_ours}
    \end{minipage}
\end{minipage}

\subsection{VGG-16 and ResNet-50 on ImageNet}
We further evaluate our method with VGG-16~\cite{Simonyan2014Very} ($13$ conv layers) and ResNet-50~\cite{HeZhaRenSun16} ($53$ conv layers) on ImageNet~\cite{DenDonSocEtAl09}. We download the open-sourced caffemodel as our pre-trained model, whose single-view top-5 accuracy on ImageNet validation dataset is $89.6\%$ (VGG-16) and $91.2\%$ (ResNet-50). For VGG-16, following SPP~\cite{Wang2018Structured}, the proportion of remaining ratios of low layers (\verb+conv1_x+ to \verb+conv3_x+), middle layers (\verb+conv4_x+) and high layers (\verb+conv5_x+) are set to $1:1.5:2$, for easy comparison. For ResNet-50, constant pruning ratio $0.4$ is adopted for all conv layers. Pruning is conducted with batch size $64$ and fixed learning rate $0.0005$, followed by retraining with batch size $256$. Experimental results are shown in Tab.\ref{tab:result_vgg16_resnet50}. On VGG-16, our method is slightly better than CP and SPP, and outperforms FP by a significant margin. Notably, since we use the same pruning ratios as SPP does, the only explanation for the performance improvement should be a better pruning process itself, guided by our incremental regularization scheme. On ResNet-50, our method is significantly better than CP and SPP, demonstrating the effectiveness of IncReg when pruning compact networks. Moreover, to confirm the actual speedup, we also evaluate our method with VGG-16 on CPU and GPU. The result is shown in Tab.\ref{tab:actual_speedup_vgg16}.

\begin{table}[!h]
    \begin{minipage}[b]{0.5\linewidth}
        \centering
        \begin{tabular}{lccc}
           \toprule
             \multirow{2}*{Method}  &  \multicolumn{3}{c}{Increased err. (\%)} \\
             \cline{2-4}
             & $2\times$ & $4\times$ & $5\times$ \\
             \hline
             TP~\cite{MolTyrKar17}  & $-$ & $4.8$ & $-$ \\
             FP~\cite{LiKadDurEtAl17} (CP's impl.) & $0.8$ & $8.6$ & $14.6$ \\
             CP~\cite{He2017Channel}              & $\mathbf{0}$ & $1.0$ & $1.7$ \\
             SPP~\cite{Wang2018Structured}             & $\mathbf{0}$  & $\mathbf{0.8}$  & $2.0$ \\
             AMC~\cite{he2018amc} & $-$ & $-$ & $\mathbf{1.4}$ \\
             Ours            & $\mathbf{0}$  & $\mathbf{0.8}$   & $1.5$ \\
           \bottomrule
        \end{tabular}
        \par\vspace{0pt}
    \end{minipage}
    \begin{minipage}[b]{0.5\linewidth}
        \centering
        \begin{tabular}{lc}
           \toprule
             Method  & Increased err. (\%) \\
             \midrule
             CP (enhanced)~\cite{He2017Channel}  & $1.4$ \\
             SPP~\cite{Wang2018Structured} & $0.8$ \\
             Ours  & $\mathbf{0.1}$ \\
           \bottomrule
        \end{tabular}
        \par\vspace{0pt}
    \end{minipage}
    \caption{Acceleration of VGG-16 (left)  and ResNet-50 (right, $2\times$ speedup) on ImageNet. The values are increased single-view top-5 error on ImageNet.}
    \label{tab:result_vgg16_resnet50}
 \end{table}

\begin{table}[!h]
    \centering
    \begin{tabular}{l p{1.5cm}<{\centering} p{1.5cm}<{\centering} p{1.5cm}<{\centering} p{1.5cm}<{\centering} p{1.5cm}<{\centering} p{1.5cm}<{\centering}}
       \toprule
         \multirow{2}*{Method}    & \multicolumn{3}{c}{CPU time (baseline: 1815 ms)}          & \multicolumn{3}{c}{GPU time (baseline: 5.159 ms)} \\
                                  \cline{2-7}
                                  & $2\times$ & $4\times$ & $5\times$                         & $2\times$ & $4\times$ & $5\times$ \\
         \hline         
         CP~\cite{He2017Channel}  & $826 (2.2\times)$ & $500 (3.6\times)$ & $449 (4.0\times)$ & $3.206 (1.6\times)$ & $2.202 (2.3\times)$ & $2.034 (2.5\times)$ \\
         Ours                     & $861 (2.1\times)$ & $469 (3.9\times)$ & $409 (4.4\times)$ & $3.225 (1.6\times)$ & $2.068 (2.5\times)$ & $1.991 (2.6\times)$ \\
       \bottomrule
    \end{tabular}
    \caption{Inference time of conv layers of CP and our method on VGG-16. Evaluation is carried out with batch size 10 and averaged by 50 runs on $224\times224$ RGB images. CPU: Intel Xeon(R) E5-2620 v4 @ 2.10GHz, single thread; GPU: GeForce GTX 1080Ti, without cuDNN. Open-sourced models of CP are used for this evaluation.}
    \label{tab:actual_speedup_vgg16}
\end{table}

\vspace{-1.5em}
\section{Conclusion}
\vspace{-0.5em}
We propose a new structured pruning method based on an incremental way of regularization, which helps CNNs to transfer their expressiveness to the rest parts during pruning by increasing the regularization factors of unimportant weight groups little by little. Our method is proved to be comparably effective on popular CNNs compared with state-of-the-art methods, especially in face of large pruning ratios and compact networks.

\textbf{Acknowledgment.} This work is supported by the Fundamental Research Funds for the Central Universities under Grant 2017FZA5007, Natural Science Foundation of Zhejiang Province under Grant LY16F010004 and Zhejiang Public Welfare Research Program under Grant 2016C31062.
\bibliographystyle{nips} 
\bibliography{nips18_v2} 

\section*{Appendix: The Proposed Theorem and Proof}
\begin{theorem}
\label{theorem:main}
Considering the objective function
\begin{equation}
E(\lambda,\omega) = L(\omega) + \frac{\lambda}{2} \omega^2,
\end{equation}
if there exists a tuple~$(\lambda_0,\omega_0)$ which satisfies the following three properties:
\begin{enumerate}
\item \label{property 1} $\lambda_0>0$;
\item \label{property 2} $L(\omega)$ has the second derivative at~$\omega_0$;
\item \label{property 3} $\omega_0$ is the local minimum of function~$Y_{\lambda_0}(\omega) = E(\lambda_0,\omega)$,
\end{enumerate}
then there exists an~$\epsilon>0$ that for any~$\lambda_1 \in (\lambda_0, \lambda_0+\epsilon)$, we can find an~$\omega_1$ which satisfies:
 \begin{enumerate}
\item $\omega_1$ is the local minimum of function~$Y_{\lambda_1}(\omega) = E(\lambda_1,\omega)$;
\item $|\omega_1| < |\omega_0|$.
\end{enumerate}
\end{theorem}

\noindent \textbf{Proof of Theorem~\ref{theorem:main}:} For a given~$\lambda>0$, the~$\omega$ which is the local minimum of the function~$Y_{\lambda}(\omega) = E(\lambda,\omega)$ should satisfy~$\frac{dY_{\lambda}(\omega)}{d\omega}=0$, which gives:
\begin{equation}
\label{eqn:function lambda omega}
\lambda = -\frac{L'(\omega)}{\omega}.
\end{equation}
In this situation, we can calculate the derivative of~$\lambda$ by using Eqn.(\ref{eqn:function lambda omega}):
\begin{equation}
\label{eqn:dlambda}
\frac{d\lambda}{d\omega} = \frac{L'(\omega)-\omega L''(\omega)}{\omega^2}.
\end{equation}
Since~$\omega_0$ is the local minimum of the function~$Y_{\lambda_0}(\omega) = E(\lambda_0,\omega)$, it should satisfy that
\begin{equation}
\left. \frac{dY_{\lambda_0}(\omega)}{d\omega} \right|_{\omega_0}=0 \textrm{ and } \left. \frac{d^2Y_{\lambda_0}(\omega)}{d\omega^2} \right|_{\omega_0}>0, \nonumber
\end{equation}
which yields
\begin{eqnarray}
\label{eqn:first}
L'(\omega_0)+\lambda_0 \omega_0&=&0 \\
\label{eqn:second}
L''(\omega_0)+\lambda_0 &>&0
\end{eqnarray}
If we take~Eqn.(\ref{eqn:first}) into~(\ref{eqn:dlambda}), we can obtain
\begin{equation}
\label{eqn:dlambda0}
\left.\frac{d\lambda}{d\omega} \right|_{(\lambda_0,\omega_0)}=-\frac{L''(\omega_0)+\lambda_0}{\omega_0}
\end{equation}

By taking~Eqn.(\ref{eqn:second}) into~(\ref{eqn:dlambda0}), we can conclude that
\begin{equation}
 \left\{
    \begin{aligned}
              & \left.\frac{d\lambda}{d\omega} \right|_{(\lambda_0,\omega_0)}<0,          &  \text{ if }  \omega_0>0\\
               & \left.\frac{d\lambda}{d\omega} \right|_{(\lambda_0,\omega_0)}>0,          &  \text{ if }  \omega_0<0\\
    \end{aligned}
    \right.
\end{equation}
In other words, when~$\omega_0$ is greater than zero, a small increment of~$\lambda_0$ will decrease the value of~$\omega_0$; and  when~$\omega_0$ is less than zero, a small increment of~$\lambda_0$ will increase the value of~$\omega_0$. In both cases, when~$\lambda_0$ increases, $|\omega_0|$ will decrease at the new local minimum of~$E(\lambda,\omega)$.

Thus, we finished the proof of Theorem~\ref{theorem:main}.

\end{document}